\documentclass[letterpaper,10pt]{article}
\usepackage[T1]{fontenc}
\usepackage{parskip}
\usepackage[font={small}]{caption}
\usepackage[hidelinks]{hyperref}
\usepackage[margin=2.5cm]{geometry}
\usepackage{times}
\usepackage{amsmath,amssymb,amsthm}
\usepackage[affil-it]{authblk}
\usepackage{listings}
\usepackage{graphicx} 
\usepackage{times} 
\usepackage{amsmath} 
\usepackage{amssymb}  
\usepackage{color}
\usepackage{caption}
\usepackage{subcaption}

\usepackage{algorithm,algorithmic,url,listings}

\newcommand{\R}{\mathbb{R}}

\newcommand{\bmat}[1]{\begin{bmatrix}#1\end{bmatrix}}

\date{}

\title{A Bayesian Filtering Algorithm for Gaussian Mixture Models}
\author[1]{Adrian G. Wills\thanks{\url{adrian.wills@newcastle.edu.au}}}
\author[2]{Johannes Hendriks\thanks{\url{johannes.n.hendriks@gmail.com}}}
\author[3]{Christopher Renton\thanks{\url{christopher.renton@newcastle.edu.au}}}
\author[4]{Brett Ninness\thanks{\url{brett.ninness@newcastle.edu.au}}}
\affil[1,2,3,4]{School of Engineering, University of Newcastle, Australia.}

\begin{document}

\maketitle

%
%
%

\begin{abstract}
  A Bayesian filtering algorithm is developed for a class of
  state-space systems that can be modelled via Gaussian mixtures. In
  general, the exact solution to this filtering problem involves an
  exponential growth in the number of mixture terms and this is
  handled here by utilising a Gaussian mixture reduction step after
  both the time and measurement updates. In addition, a
  square-root implementation of the unified algorithm is presented and
  this algorithm is profiled on several simulated systems. This
  includes the state estimation for two non-linear systems that are
  strictly outside the class considered in this paper.
\end{abstract}



\section{INTRODUCTION}
The problem of estimating the state of a system based on noisy
observations of the system outputs has received significant research
attention for more than half a century~\cite{RisticAG:2004}. This
attention stems from the fact that state estimation is used in many
areas of science and engineering, including---for example---guidance,
navigation and control of autonomous
vehicles~\cite{thrun2005probabilistic}, target
tracking~\cite{ardeshiri2012mixture}, fault
diagnosis~\cite{yu2012particle}, system identification~\cite{SPM675}
and many other related areas.

This research has resulted in many approaches to the
state-estimation problem including the much celebrated Kalman
filter~\cite{KalmanFilter}, extended Kalman filter~\cite{smith1962},
Unscented Kalman filter~\cite{julier1995new} and Sequential
Monte-Carlo (SMC) approaches~\cite{Gordon:1993}. Each of these variants
exploits different structural elements of the state-space model and
each has known strengths and associated weaknesses. For example, if
the system is linear and Gaussian, then the Kalman filter is the most
obvious choice. On the other hand, if the system is highly
non-linear then SMC methods may be the most suitable choice.

In this paper, we consider state estimation for a class of state-space
models that can be described by Gaussian mixture models, for both the
process and measurement models. This model class captures a broad
range of systems including, for example, stochastically switched
linear Gaussian systems, systems that exhibit multi-modal state and/or
measurement noise, and systems that exhibit long--tailed stochastic
behaviour, to name a few. In theory, Gaussian mixtures are suitable
for modelling a large class of probability
distributions~\cite{bacharoglou2010approximation}.

Inclusion of Gaussian mixtures in the Bayesian filtering framework
dates at least back the work in~\cite{alspach1972nonlinear}, where a
mixture was employed to represent the predicted and filtered densities
for general non-linear systems. As identified
in~\cite{alspach1972nonlinear}, a significant drawback of this
approach is that the number of mixture components can grow rapidly as
the filter progresses. Subsequent work in this area has concentrated
efforts towards ameliorating this problem by reducing the
mixture. This includes approaches based on SMC methods, Expectation
Maximisation (EM) clustering, Unscented transforms and many related methods~\cite{simandl2005sigma,kotecha2003gaussian,raihan2016particle,fearnhead2004particle}. A
common theme among these contributions is that they employ particle
resampling techniques to reduce the number of components that need to
be tracked. This is either achieved directly by removing highly
unlikely components of the mixture, or numerically via resampling
algorithms.

Gaussian mixtures are also employed within the area of multiple
target tracking problems~\cite{ardeshiri2012mixture}. Within this field, the
methods of Multiple-Hypothesis Kalman Trackers (MHT) and Gaussian Mixture
Probability Hypothesis Density filters (GM-PHD) rely on similar ideas
presented here, albeit for the target tracking
problem~\cite{ardeshiri2012mixture}. The key idea here is remove
unlikely targets from the list of possible targets using a pruning
mechanism. 

Closely related to these ideas is the independent area of Gaussian
mixture model reduction. The recent review in \cite{crouse2011}
compares several of the main contenders in this field. The methods are
compared based on accuracy of the reduced mixture relative to the
original one, and the efficiency of each algorithm. The conclusion
notes that the Kullback--Leibler reduction method
of~\cite{runnalls2007kullback} is both efficient and appears to
perform well in terms of accurately reducing the Gaussian mixture.

The contribution of this paper is to unite the GMM state-space model
structure together with the Kullback-Leibler reduction algorithm to
deliver a new Bayesian filtering algorithm for state estimation of GMM state-space
models. An important aspect of this algorithm is the numerically
stable and efficient implementation by propagating covariance
information in square-root form. This relies on the novel contribution
of a square-root form Kullback-Leibler GMM reduction algorithm. 

\section{PROBLEM DESCRIPTION}
Consider a general state-space model expressed via
a state transition probability and measurement likelihood
\begin{align}
  x_{t+1} &\sim p(x_{t+1} | x_t),  \label{eq:1}\\
y_t &\sim p(y_t | x_t), \label{eq:2}
\end{align}
where the state $x_t \in \R^n$ and the output $y_t \in
\R^p$. The state transition probability distribution \eqref{eq:1} and
the likelihood \eqref{eq:2} may be parameter dependent and may
also be time-varying, although these embellishments have been ignored
for ease of exposition. 

Given a collection of observations $Y_N = \{y_1, \ldots,y_N\}$, then
the general Bayesian filtering problem can be solved recursively via
the well known time and measurement equations
\begin{align}
  \label{eq:3}
  p(x_{t} | Y_t) &= \frac{p(y_t | x_t)\,p(x_t | Y_{t-1})}{p(y_t |
                   Y_{t-1})}, \\
  p(x_{t+1} | Y_t) &= \int p(x_{t+1} | x_t) \, p(x_t | Y_t) \, dx_t.\label{eq:4}
\end{align}
Solving equations \eqref{eq:3}--\eqref{eq:4} has attracted enormous
attention for many decades and this general approach has been
successfully employed across disparate areas of science and
engineering. In the general case, the solutions to
equations \eqref{eq:3} and \eqref{eq:4} cannot be expressed in closed form, and this has been the focus of significant research
activity. Indeed, this difficulty has led to many approximation
methods including the employment of Extended Kalman Filters, Unscented
Kalman Filters and Sequential Monte-Carlo (Particle) Filters to name
but a few.

In this paper, we restrict our attention to a class of
state-space systems that are not as general as
\eqref{eq:1}--\eqref{eq:2}, yet---in theory---have the potential to
approximate general state-space systems arbitrarily
well\cite{bacharoglou2010approximation}. More precisely, we will
consider state-space systems that can be described by a Gaussian
Mixture Model (GMM) in the following manner. The prior is described by
\begin{equation}
  \label{eq:5}
  p(x_1) = \sum_{i=1}^{N_p} \alpha_i \,
  \mathcal{N}(x_1;\,\mu_1^i,\,P_1^i),\quad \sum_{i=1}^{N_p} \alpha_i = 1.
\end{equation}
The process model is of the form
\begin{equation}
  \label{eq:6}
  p(x_{t+1} | x_t) = \sum_{j=1}^{N_x} \beta_t^j \, \mathcal{N}(x_{t+1} ;\,
  A_t^j x_t + u_t^j,\,Q_t^j),\quad \sum_{j=1}^{N_x} \beta_t^j = 1.
\end{equation}
The measurement model is of the form
\begin{equation}
  \label{eq:7}
  p(y_t | x_t) =
  \sum_{k=1}^{N_y}\gamma_t^k \, \mathcal{N}(y_t;\,
  C_t^kx_t+v_t^k,\, R_t^k),\quad\sum_{k=1}^{N_y}\gamma_t^k = 1.
\end{equation}
In the above, the form $\mathcal{N}(x;\,\mu,\, P)$ is used to denote a
standard multivariate Gaussian distribution. The mean offset terms
$u_t^j$ and $v_t^k$ allow for the inclusion of input signals or other generated
signals that do not depend on $x_t$. 

The model in \eqref{eq:5}--\eqref{eq:7} is not as general as
\eqref{eq:1}--\eqref{eq:2}. However, the primary advantage of
restricting attention to this Gaussian mixture model class is that the
time and measurement update equations can be expressed in closed
form. Indeed, this fact has already been exploited by many authors
dating back to the work of~\cite{alspach1972nonlinear}. Therein, the
Authors raise a serious issue with this approach, which can be observed by
following the recursions in \eqref{eq:3}--\eqref{eq:4} for just a few
steps. Specifically, starting with the prior \eqref{eq:5}, then the
first measurement update can be expressed as {\small
\begin{align}
  \label{eq:8}
    p(x_{1} | Y_1) &= \sum_{i=1}^{Np} \sum_{k=1}^{N_y} \alpha_i
                     \gamma_k
                     \frac{\mathcal{N}(y_1;
  C_1^kx_1+v_1^k,R_1^k) \, \mathcal{N}(x_1;\mu_1^i,P_1^i)}{p(y_1 |
                   Y_{0})},
\end{align}
}
and updating this filtered state distribution to the predicted state
via \eqref{eq:4} results in
\begin{align}
  \label{eq:9}
  p(x_{2} | Y_1) &= \sum_{i=1}^{N_p} \sum_{j=1}^{N_x} \sum_{k=1}^{N_y} \alpha_i
                     \beta_1^j \gamma_1^k\int f_1^{i,j,k}(x_2,x_1,Y_1) \,dx_1,
\end{align}
where
\begin{align}
  \label{eq:10}
  f_1^{i,j,k}(x_2,x_1,Y_1) &\triangleq \frac{\mathcal{N}(y_1;
  C_1^kx_1+v_1^k,R_1^k) }{p(y_1 |
                   Y_{0})} \nonumber\\
  &\quad \times \mathcal{N}(x_{2} ;
  A_1^j x_1 + u_1^j,Q_1^j) \, \mathcal{N}(x_1;\mu_1^i,P_1^i).
\end{align}
Therefore, the predicted state distribution is already a GMM with
$N_p N_x N_y$ components. This becomes unmanageable whenever either $N_x$ or $N_y$ are greater than 1, and the number of
measurements $N$ becomes large. For example, if $N_p = 1$, $N_x = 1$
and $N_y=2$, then a very modest $N=100$ measurements would result in
the predicted state density composed of
$N_y^{100} = 2^{100} \approx 10^{30}$ Gaussian components---of the
same order as the estimated number of bacterial cells on Earth
\cite{whitman1998prokaryotes}. Clearly this is not practical and the
Authors in~\cite{alspach1972nonlinear} suggest that the number of
terms should be reduced after each iteration of the filter, but do not
provide a suitable mechanism for achieving this.


In the current paper, we adopt the approach of maintaining a
prediction and filtering Gaussian mixture and utilise the work in
\cite{runnalls2007kullback} to reduce the mixture at each stage via a
Kullback-Leibler discrimination approach. Akin to resampling, this
approach compresses the distribution while maintaining the most
prominent aspects of the mixture model. This approach will be outlined
in the following section.

\section{FILTERING ALGORITHM}
\subsection{A Kullback-Leibler GMM Reduction Method}
In this section, we will outline the important aspects of a
Kullback-Leibler Gaussian Mixture Model reduction method proposed
in~\cite{runnalls2007kullback} and show how this can be utilised
within a Bayesian filtering framework in the following
subsection. Importantly, we extend the work
in~\cite{runnalls2007kullback} in a trivial manner to allow for a
reduction of the GMM based on a user-defined threshold, and this has
the effect of adapting the mixture to a maximum acceptable loss of
information.

The main idea in~\cite{runnalls2007kullback} is to reduce a GMM
\begin{align}
  \label{eq:11}
  \pi(x) = \sum_{i=1}^N w_i \pi_i(x), \quad \pi_i = \mathcal{N}(x;
  \mu_i, P_i), \quad \sum_{i=1}^N w_i = 1,
\end{align}
to another mixture model
\begin{align}
  \label{eq:12}
  \eta(x) = \sum_{i=1}^M v_i \eta_i(x), \quad \pi_i = \mathcal{N}(x;
  \nu_i, Q_i), \quad \sum_{i=1}^M v_i = 1,
\end{align}
where $1 \leq M \leq N$ and in this subsection the variables $x$, $P$
and $Q$ and others are to be treated as general variables and are not
meant to represent states and covariance matrices defined elsewhere.

The mechanism proposed in~\cite{runnalls2007kullback} to achieve this reduction is to form a
component of $\eta(x)$ by merging two components from $\pi(x)$, that
is,
\begin{align}
  \label{eq:13}
  \eta_k(x) = f(\pi_i(x),\pi_j(x)).
\end{align}
This is repeated until the desired number of components $M$ is
achieved.  How to merge these two components from $\pi(x)$ and which
two to choose are detailed in~\cite{runnalls2007kullback}, but here we
present the salient features. In terms of the merging function
$f(\cdot,\cdot)$, the Authors in \cite{runnalls2007kullback} employ
the following merge that preserves the first- and second-order
moments of the original two components
\begin{align}
  \label{eq:14}
  f(\pi_i(x),\pi_j(x)) = w_{ij}\,\mathcal{N}(x; \mu_{ij} , P_{ij}),
\end{align}
where
\begin{align}
  \label{eq:15}
  w_{ij} &= w_i + w_j, \\
  \mu_{ij} &= w_{i|ij}\mu_i + w_{j|ij} \mu_j\label{eq:16},\\
  P_{ij} &= w_{i|ij}P_i + w_{j|ij} P_j\nonumber \\ &\qquad+ w_{i|ij}
           w_{j|ij}(\mu_i-\mu_j)(\mu_i-\mu_j)^T,\label{eq:17}\\
  w_{i|ij} &= \frac{w_i}{w_i + w_j},\label{eq:18}\\
  w_{j|ij} &= \frac{w_j}{w_i + w_j}.\label{eq:19}
\end{align}
It is possible (see Section~IV in~\cite{runnalls2007kullback}) to
bound the Kullback-Leibler discrimination between $\eta(x)$ and
$\pi(x)$ for each reduction. This is not a bound on the difference
between $\eta(x)$ and $\pi(x)$, but provides a bound on the discrimination
between the mixtures before and after reduction of a component pair. This
bound is denoted $B(i,j)$ and is defined as
\begin{align}
  \label{eq:20}
  B(i,j) &\triangleq \frac{1}{2}\Bigl [ w_{ij} \log |P_{ij}| - w_i
           \log |P_i| - w_j \log |P_j| \Bigr ].
\end{align}
In the above, we have used the notation $|\cdot|$ to represent the
matrix determinant. 

The utility of the bound $B(i,j)$ is that we can choose among the
possible $i$'s and $j$'s to find the combination that generates the
smallest value $B(i,j)$, which therefore represents a bound on the
smallest Kullback-Leibler discrimination among the possible
components. That is, merging the $(i,j)$ mixture components results in
the smallest change to the mixture according to the Kullback-Leibler
discrimination bound. Importantly, $B(i,j) = B(j,i)$ and $B(i,i) = 0$
so it is only necessary to search
$\frac12 N (N-1)$
combinations---less than half of all possible $(i,j)$ pairs.

Once the minimising $(i,j)$-pair is found, then the
corresponding components can be merged and the process can be repeated
on the new, reduced, mixture. It is worth noting that this approach is
completely deterministic in that the resulting data
$\{v_i, \nu_i, Q_i\}_{i=1}^M$ that describes the reduced mixture is
only a function the original data $\{w_i, \mu_i, P_i\}_{i=1}^N$ that
describes the starting mixture.

In the current paper, we are less concerned with \emph{a priori}
fixing the number $M$ of components in the reduced mixture, and are
more concerned with minimising the number of components at each
iteration of the filter. To that end, here we will detail an algorithm
that continues to merge components until the bound $B(i,j)$ exceeds a
given threshold. This may be related back to an acceptable loss of
accuracy and, although not explored here, it may be possible to
compensate for this decrease in entropy by increasing the
corresponding covariance terms. This aside, Algorithm~\ref{alg:KLR}
produces a new mixture by successive merging until a threshold in
$B(i,j)$ is exceeded, or, a desired minimum number of components is
achieved.

\begin{algorithm}[!t]
\caption{\textsf{Kullback-Leibler GMM Reduction}}
\small
\begin{algorithmic}[1]
  \REQUIRE Integers $M_l>0$ and $M_u>0$ that determine a minimum and
  maximum, respectively, number of components in the reduced mixture,
  and a threshold $\lambda > 0$, and the initial mixture data
  $\{w_i, \mu_i, P_i\}_{i=1}^N$.
  \STATE Set $k=N$ and define a set of integers
  $S \triangleq \{1,\dotsc,N\}$. %
  \STATE Calculate the bound matrix
  entries $B(i,j)$ using \eqref{eq:20} for $i \in S$ and $j \in S$
  noting that $B(i,i) = 0$ and $B(i,j) = B(j,i)$.
\WHILE{$k > M_u$ \OR ($k > M_l$ \AND $\min_{i,j}B(i,j) < \lambda$)}
  \STATE Find $(i^\star,j^\star) = \arg \min_{i,j}B(i,j)$.
  \STATE Merge components $(i^\star,j^\star)$ using \eqref{eq:15}--\eqref{eq:19} so that the
  $i^\star$'th component data is replaced with
  \begin{align}
    \label{eq:21}
    \{w_{i^\star}, \mu_{i^\star}, P_{i^\star} \} \leftarrow \{
    w_{i^\star j^\star}, \mu_{i^\star j^\star}, P_{i^\star j^\star} \}.
  \end{align}
  \STATE Remove $j^\star$ from the set $S$ so that $S \leftarrow S \setminus
  j^\star$.
  \STATE Set $k \leftarrow k -1$.
  \STATE Re-calculate the bound matrix
  entries $B(i,j)$ using \eqref{eq:20} for $i \in S$ and $j \in S$
  noting that $B(i,i) = 0$ and $B(i,j) = B(j,i)$ and noting that only
  the $i^\star$'th row and column will have changed.
  \ENDWHILE
\end{algorithmic}
\label{alg:KLR}
\end{algorithm}

\subsection{Filtering Algorithm}
In this section we will detail the combination of the Bayesian
filtering recursions \eqref{eq:3}--\eqref{eq:4} for the class of
Gaussian mixture state-space models defined in
\eqref{eq:5}--\eqref{eq:7}, where the number of components in the
filtering and predicted mixtures will be reduced at each stage by
utilising Algorithm~\ref{alg:KLR}. Recall that the reduction is
necessary in order to combat the exponential growth in the number of
mixture components.

To this end, assume that we have available a prediction mixture given
by
{\small
\begin{align}
  \label{eq:24}
  p(x_{t} | Y_{t-1}) &= \sum_{\ell = 1}^{N_{t|t-1}} w_{t|t-1}^\ell \mathcal{N}(x_t;
                   \widehat{x}_{t|t-1}^\ell, P_{t|t-1}^\ell ),\\
                     &\qquad \qquad                   \sum_{\ell=1}^{N_{t|t-1}} w_{t|t-1}^\ell = 1. \nonumber
\end{align}
}%
Note that the prior \eqref{eq:5} is already in this form at
$t=1$. Consider the measurement update \eqref{eq:3}, which can be
expressed in the current GMM setting as {\small
\begin{align}
 \label{eq:22}    
  p(x_{t} | Y_t) &= \sum_{\ell=1}^{N_{t|t-1}} \sum_{k=1}^{N_y} w_{t|t-1}^\ell
                     \gamma_t^k
                     \frac{\mathcal{N}(y_t;
  C_t^kx_t+v_t^k,R_t^k) }{p(y_t |
                   Y_{t-1})}\nonumber\\
  &\qquad \qquad \qquad \qquad \times \mathcal{N}(x_t;\widehat{x}_{t|t-1}^\ell , P_{t|t-1}^\ell).
\end{align}
}
Due to the linear Gaussian structure, this can be expressed as (see
Section III in~\cite{alspach1972nonlinear})
\begin{align}
  \label{eq:23}
  p(x_{t} | Y_t) &= \sum_{s = 1}^{N_{t|t}} w_{t|t}^s \,\mathcal{N}(x_t;
                   \widehat{x}_{t|t}^s, P_{t|t}^s ), \qquad
                   \sum_{s=1}^{N_{t|t}} w_{t|t}^s = 1,
\end{align}
where for each $k=1,\ldots,N_y$ and  $\ell = 1,\ldots,N_{t|t-1}$ it
holds that
\begin{align}
  \label{eq:25}
  N_{t|t} &= N_{t|t-1}N_y,\\
  s &\triangleq N_y(\ell - 1) + k,\label{eq:26}\\
  \widehat{x}_{t|t}^s &= \widehat{x}_{t|t-1}^\ell + K_t^s e_t^s,\\
  e_t^s &= y_t - C_t^k\widehat{x}_{t|t-1}^\ell - v_t^k,
\end{align}
and
\begin{align}
  \Sigma_t^s &= C_t^kP_{t|t-1}^\ell (C_t^k)^T + R_t^k,\\
  K_t^s &= P_{t|t-1}^\ell C_t^k (\Sigma_t^s)^{-1},\\
  P_{t|t}^s &= P_{t|t-1}^\ell -   K_t^s \Sigma_t^s (K_t^s)^T,\\
  w_{t|t}^s &=
              \frac{\bar{w}_{t|t}^s}{\sum_{s=1}^{N_{t|t}}\bar{w}_{t|t}^s},\\
  \bar{w}_{t|t}^s &= \frac{w_{t|t-1}^\ell \gamma_t^k \exp(-\frac{1}{2}(e_t^s)^T(\Sigma_t^s)^{-1}e_t^s)}{(2\pi)^{n/2}
                    |\Sigma_t^s|^{1/2}}. \label{eq:31}
\end{align}
With this filtering distribution in place, then we can proceed with the
time update \eqref{eq:4}, which can be expressed as
{\small
\begin{align}
  \label{eq:27}
  p(x_{t+1} | Y_t) &= \sum_{s=1}^{N_{t|t}} \sum_{j=1}^{N_x} w_{t|t}^s
                     \beta_t^j \nonumber \\
  &\quad \times \int
                     \mathcal{N}(x_{t+1}; A_t^j x_t + u_t^j, Q_t^j) \, \mathcal{N}(x_t;
                   \widehat{x}_{t|t}^s, P_{t|t}^s ) \,dx_t.
\end{align}
}
Again, due to the linear Gaussian densities involved, we can express
this predicted mixture via
\begin{align}
  \label{eq:28}
  p(x_{t+1} | Y_{t}) &= \sum_{\ell = 1}^{N_{t+1|t}} w_{t+1|t}^\ell \,\mathcal{N}(x_{t+1};
                   \widehat{x}_{t+1|t}^\ell, P_{t+1|t}^\ell ),\\
                     &\qquad \qquad                   \sum_{\ell=1}^{N_{t+1|t}} w_{t+1|t}^\ell = 1, \nonumber
\end{align}
where for each $s = 1,\ldots,N_{t|t}$ and $j=1,\ldots,N_x$ we have that
\begin{align}
  \label{eq:29}
  N_{t+1|t} &= N_{t|t}N_x,\\
  \ell &\triangleq N_x(s-1) + j,\\
  \widehat{x}_{t+1|t} &= A_t^j \widehat{x}_{t|t}^s + u_t^j,\\
  P_{t+1|t} &= A_t^j P_{t|t} (A_t^j)^T + Q_t^j,\\
  w_{t+1|t}^\ell &= w_{t|t}^s \beta_t^j.\label{eq:32}
\end{align}
Therefore, we have arrived back at our initial
starting assumption in \eqref{eq:24}, albeit one time step
ahead. Hence, the recursion can repeat.

What remains is to reduce the filtered mixture $p(x_t | Y_t)$ and the
predicted mixture $p(x_{t+1} | Y_t)$ at each iteration in order to
avoid exponential growth in computational load. To this end, below we
define an algorithm that combines the above filtering recursions with
Algorithm~\ref{alg:KLR} to provide a practical Bayesian Filtering
algorithm for Gaussian Mixture Model state-space systems.
\begin{algorithm}[!t]
\caption{\textsf{GMM Filter}}
\small
\begin{algorithmic}[1]
  \REQUIRE Integers $M_{fl} >0, M_{fu}>0, M_{pl}>0, M_{pu}>0$ that
  determine the minimum and maximum number of components in the
  filtering and prediction mixtures after reduction, and threshold
  values $\lambda_f > 0$ and $\lambda_p > 0$ that represent the
  filtering and prediction thresholds, respectively, for
  Algorithm~\ref{alg:KLR}.  
  \STATE Set $t=1$ and define the initial
  prediction mixture at $t=1$ according to the prior \eqref{eq:5} so
  that for each $i = 1,\ldots,N_p$
  \begin{align}
    \label{eq:30}
    N_{1|0} \triangleq N_p,\quad
    w_{1|0}^i \triangleq \alpha_i,\quad
    \widehat{x}_{1|0}^i \triangleq \mu_1^i,\quad
    P_{1|0}^i \triangleq P_1^i.
  \end{align}
\WHILE{$k \leq N$}
  \STATE Calculate $p(x_t | Y_t)$ according to
  \eqref{eq:23} and \eqref{eq:25}--\eqref{eq:31}.
  \STATE Replace $p(x_t | Y_t)$ with a reduced mixture using
  Algorithm~\ref{alg:KLR} with $M_l=M_{fl}$ and $M_u=M_{fu}$, $\lambda=\lambda_f$ and where
  the mixture data corresponds to $\{w_{t|t}^s, \widehat{x}_{t|t}^s,
  P_{t|t}^s\}_{s=1}^{N_{t|t}}$.
  \STATE Calculate the prediction mixture $p(x_{t+1} | Y_t)$ according
  to \eqref{eq:28} and \eqref{eq:29}--\eqref{eq:32}.
  \STATE Replace $p(x_{t+1} | Y_t)$ using
  Algorithm~\ref{alg:KLR} with $M_l=M_{pl}$ and $M_u=M_{pu}$, $\lambda=\lambda_p$ and where
  the mixture data corresponds to $\{w_{t+1|t}^\ell, \widehat{x}_{t+1|t}^\ell,
  P_{t+1|t}^\ell\}_{\ell=1}^{N_{t+1|t}}$.
  \ENDWHILE
\end{algorithmic}
\label{alg:GMMF}
\end{algorithm}

\section{NUMERICALLY STABLE IMPLEMENTATION}
The above discussion outlines Algorithm~\ref{alg:GMMF}, which produces
estimates of the filtering and prediction mixture densities. The main
computational tools employed are those common to Kalman Filtering and
those introduced in Algorithm~\ref{alg:KLR} to perform the required
mixture reduction. 

It is well known~\cite{KAILATHNEW} that when implementing Kalman
Filters, care should be taken in ensuring that the covariance matrices
remain positive definite and symmetric. Unfortunately, the equations in
\eqref{eq:25}--\eqref{eq:31} and \eqref{eq:29}--\eqref{eq:32} are not
guaranteed to maintain this requirement if implemented naively. To
circumvent this problem, we have employed a square-root version of the
filtering recursions. 

We present the essential idea here in order to help explain a
square-root implementation of Algorithm~\ref{alg:KLR} that relies on
the mixture being provided in square-root form. To this end, it is assumed
that we have access to square-root versions of the covariance matrices
\begin{align}
  \label{eq:33}
  P_1^i &= (P_1^i)^{T/2} (P_1^i)^{1/2},\\
  Q_t^j &= (Q_t^j)^{T/2} (Q_t^j)^{1/2}, \\
  R_t^k &= (R_t^k)^{T/2} (R_t^k)^{1/2}.
\end{align}
The measurement update equations \eqref{eq:25}--\eqref{eq:31} can then
be calculated by employing a QR factorisation as follows
\begin{align}
  \bmat{\mathcal{R}_{11}^s & \mathcal{R}_{12}^s \\ 0 & \mathcal{R}_{22}^s } &\triangleq
                                                \mathcal{Q}^s
                                                \bmat{(R_t^k)^{1/2} & 0 \\
  (P_{t|t-1}^\ell)^{1/2} (C_t^k)^T & (P_{t|t-1}^\ell)^{1/2}}.
\end{align}
In the above $\mathcal{Q}^s$ is an orthonormal matrix and
$\mathcal{R}_{11}^s$ and $\mathcal{R}_{22}^s$ are upper triangular
matrices. Note that we have dropped the time reference from these
matrices for ease of exposition. This then allows for the computation
of the remaining terms via
\begin{align}
  (\Sigma_t^s)^{1/2} &= \mathcal{R}_{11}^s,\\
  (P_{t|t}^s)^{1/2} &= \mathcal{R}_{22}^s,\\
   \tilde{e}_t &= (\mathcal{R}_{11}^s)^{-T} \left (y_t -
                 C_t^k\widehat{x}_{t|t-1}^\ell - v_t^k \right ),\\
  \widehat{x}_{t|t}^s &= \widehat{x}_{t|t-1}^\ell +
                        (\mathcal{R}_{12}^s)^T  \tilde{e}_t^s,
\end{align}
and the weights can be calculated via
\begin{align}
  w_{t|t}^s &=
              \frac{\bar{w}_{t|t}^s}{\sum_{s=1}^{N_{t|t}}\bar{w}_{t|t}^s},\\
  \bar{w}_{t|t}^s &= \frac{w_{t|t-1}^\ell \gamma_t^k \exp(-\frac{1}{2}(\tilde{e}_t^s)^T\tilde{e}_t^s)}{(2\pi)^{n/2}
                    \sigma_t^s},\\
  \sigma_t^s &\triangleq \left ( \prod_{i=1}^p |\Sigma_t^s(i,i)|^2 \right )^{1/2}.
\end{align}
Similar arguments can be employed in the prediction step with
\begin{align}
  \label{eq:35}
  \bmat{\bar{\mathcal{R}}^\ell\\0} &\triangleq \bar{\mathcal{Q}}^\ell
                                     \bmat{ (P_{t|t}^s)^{1/2}
                                     (A_t^j)^T \\ (Q_t^j)^{1/2}},\\
  (P_{t+1|t}^\ell)^{1/2} &= \bar{\mathcal{R}}^\ell.
\end{align}
The remaining prediction equations in \eqref{eq:29}--\eqref{eq:32}
are unchanged.

Therefore, we can compute all the required filtering and prediction
covariances in square-root form. Aside from the numerical stability
that this brings, we can also exploit the square-root form in
Algorithm~\ref{alg:KLR}. Specifically, referring to the steps in
Algorithm~\ref{alg:KLR}, it is important that we can compute the bound
$B(i,j)$ efficiently and robustly. To this end, note that according to
\eqref{eq:15}--\eqref{eq:19}, if the covariance matrices
$\{P_i, P_j\}$ are provided in square-root form then we can compute
the following QR factorisation
\begin{align}
  \label{eq:37}
  \bmat{\tilde{\mathcal{R}} \\ 0} &= \tilde{\mathcal{Q}} \bmat{\sqrt{w_{i|ij}} P_i^{1/2} \\ \sqrt{w_{j|ij}}P_j^{1/2}
  \\ \sqrt{w_{i|ij}w_{j|ij}}(\mu_i - \mu_j)^T},
\end{align}
so that 
{\small
\begin{align}
  \label{eq:38}
  \tilde{\mathcal{R}}^T \tilde{\mathcal{R}} &= \bmat{\sqrt{w_{i|ij}} P_i^{1/2} \\ \sqrt{w_{j|ij}}P_j^{1/2}
  \\ \sqrt{w_{i|ij}w_{j|ij}}(\mu_i - \mu_j)^T} ^T \bmat{\sqrt{w_{i|ij}} P_i^{1/2} \\ \sqrt{w_{j|ij}}P_j^{1/2}
  \\ \sqrt{w_{i|ij}w_{j|ij}}(\mu_i - \mu_j)^T}\nonumber \\
                                            &= w_{i|ij}P_i^{T/2}P_i +
                                              w_{j|ij}P_j^{T/2}P_j^{1/2}\nonumber
  \\ &\qquad  \qquad + w_{i|ij}w_{j|ij}(\mu_i - \mu_j) (\mu_i - \mu_j)^T\nonumber\\
                                            &= P_{ij}.
\end{align}
}
Therefore,
\begin{align}
  \label{eq:39}
  P_{ij}^{1/2} &= \tilde{\mathcal{R}}
\end{align}
is a square-root factor for the merged component. In addition, the
bound $B(i,j)$ can be readily calculated by exploiting the fact that
for any upper triangular matrix $A \in \R^{m \times m}$ 
\begin{align}
  \label{eq:41}
  \log \left|A^TA \right| = 2 \sum_{i=1}^m \log |A(i,i)|
\end{align}
to deliver
{\small
\begin{align}
  \label{eq:40}
  B(i,j) &= \sum_{i=1}^n w_{ij}\log |P_{ij}^{1/2}(i,i)| - w_i\log |P_i^{1/2}| -
           w_j\log |P_j^{1/2}|.
\end{align}
}
Therefore, we have shown that the square-root form of the Kalman
filter covariance matrices will be maintained after calling the
mixture reduction algorithm by employing \eqref{eq:37} and
\eqref{eq:39}. We have further shown how to readily compute $B(i,j)$
that exploits this square-root form.

The QR factorisation involved in \eqref{eq:37} can also exploit the
special sparse structure arising from two stacked upper triangular
matrices and one row vector. This feature has been exploited in our
implementation, but not detailed here further.
\section{EXAMPLES}
In this section, we present a collection of examples that profile the
proposed Algorithm~\ref{alg:GMMF}, henceforth referred to as the GMMF
approach. We move from a simple linear state-space model with additive
Gaussian noise, through to a non-linear and time-varying system that
falls outside the modelling assumptions in this paper.

\subsection{Linear State-Space Model}
In this example, we profile Algorithm~\ref{alg:GMMF} on a
well known and studied problem of filtering for linear state-space
systems with additive Gaussian noise on both the state and
measurements. Clearly this falls within our model assumptions
\eqref{eq:5}--\eqref{eq:7} and our purpose here is to observe that
even if we deliberately start with more mixture components in the
prior $p(x_1)$ than are strictly necessary, the algorithm will very
quickly reduce the number of mixture terms in both the predicted and
filtered densities.

\begin{figure}[!bth]
  \includegraphics[width=\columnwidth]{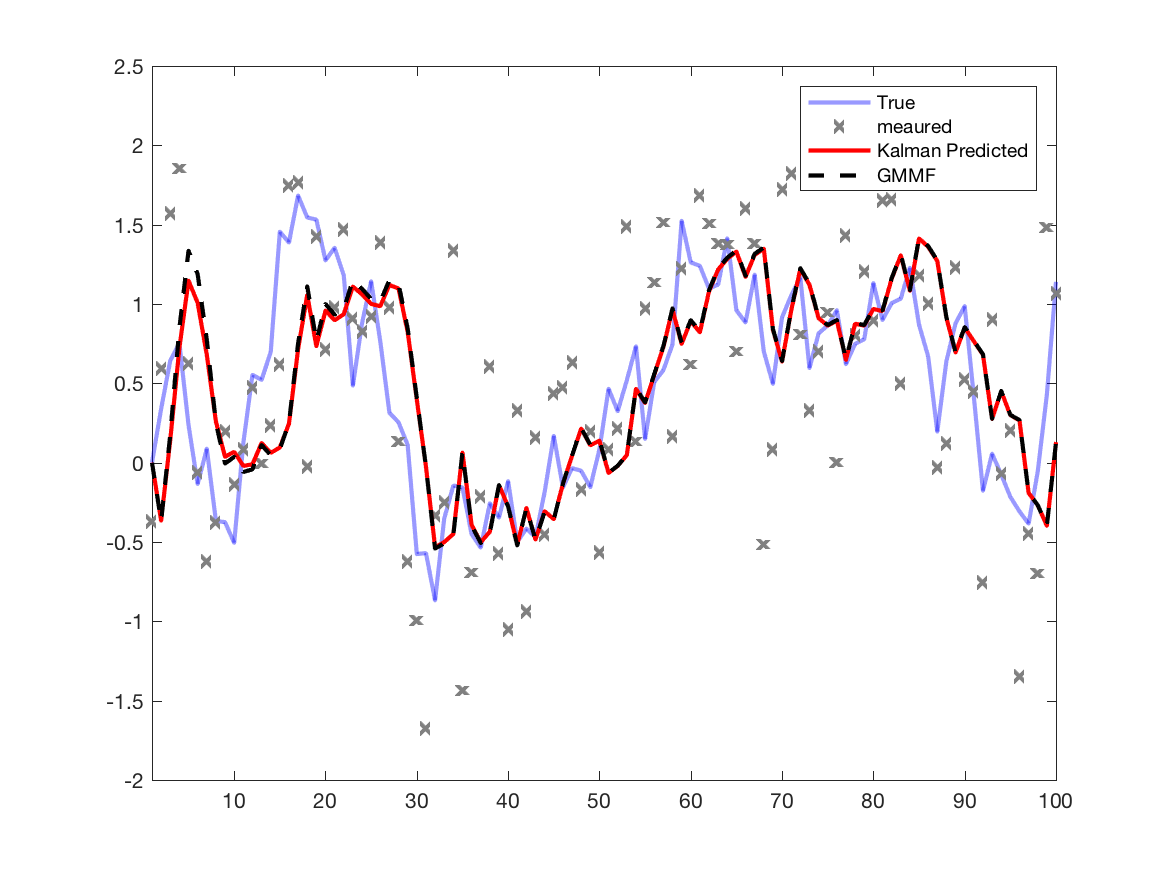}
  \caption{Plot of the predicted mean for the first state from both
    the Kalman Filter (red) and GMM Filter (dashed black) against the true
    state (solid blue). Measurements are indicated as crosses.}
  \label{fig:linear_time}
\end{figure}

\begin{figure}[!bth]
    \centering
    \begin{subfigure}[t]{.49\linewidth}
        \includegraphics[width=\columnwidth]{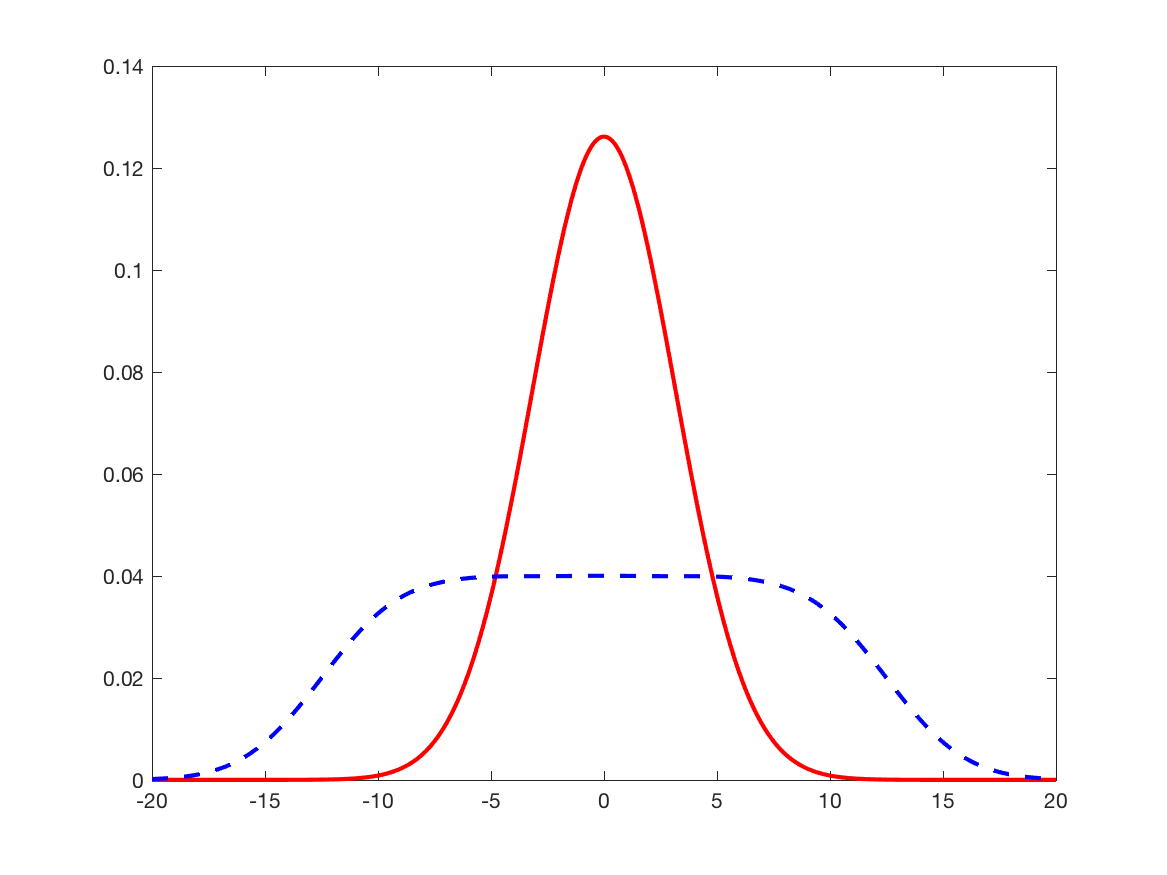}
       \caption{Time $t=1$.}
        \label{fig:lin1}
      \end{subfigure}
      \begin{subfigure}[t]{.49\linewidth}
        \includegraphics[width=\columnwidth]{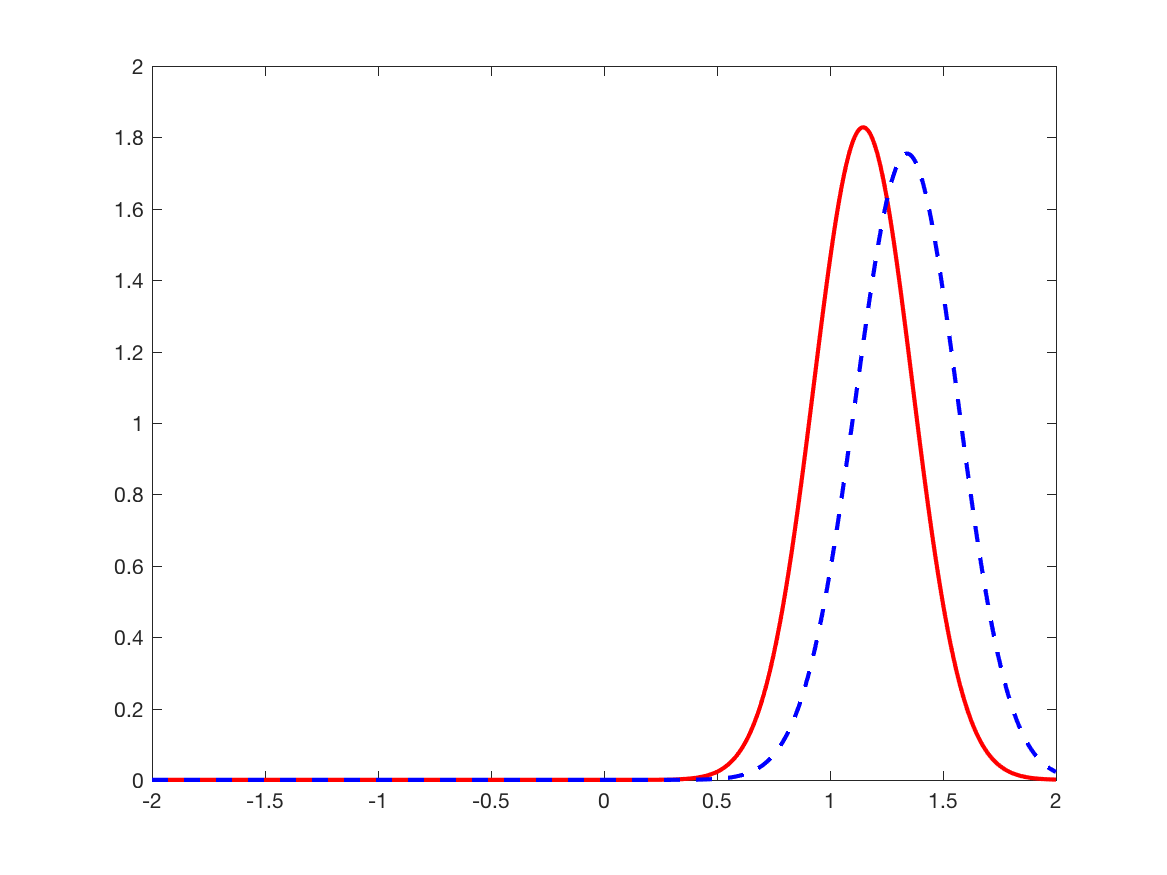}
        \caption{Time $t=5$.}
        \label{fig:lin2}
      \end{subfigure}
      \begin{subfigure}[t]{.49\linewidth}
        \includegraphics[width=\columnwidth]{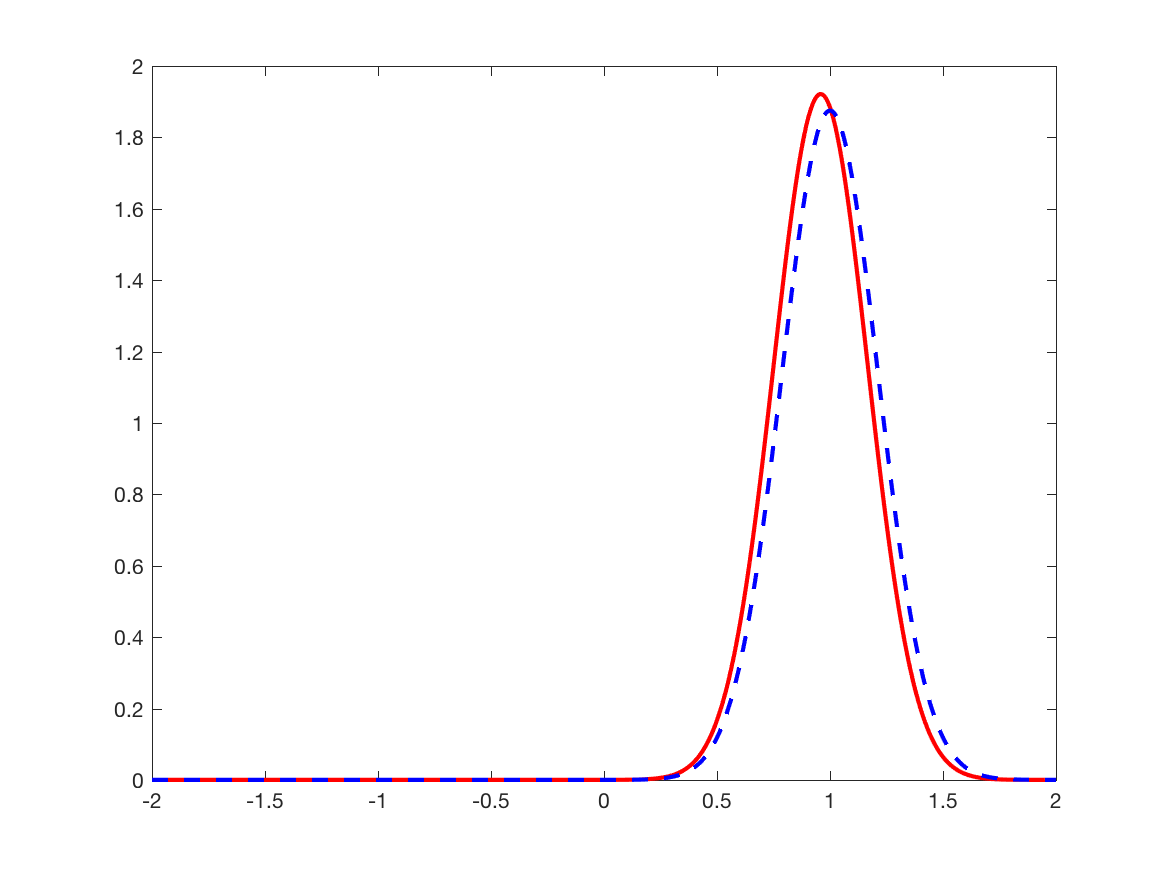}
       \caption{Time $t=20$.}
        \label{fig:lin3}
      \end{subfigure}
      \begin{subfigure}[t]{.49\linewidth}
        \includegraphics[width=\columnwidth]{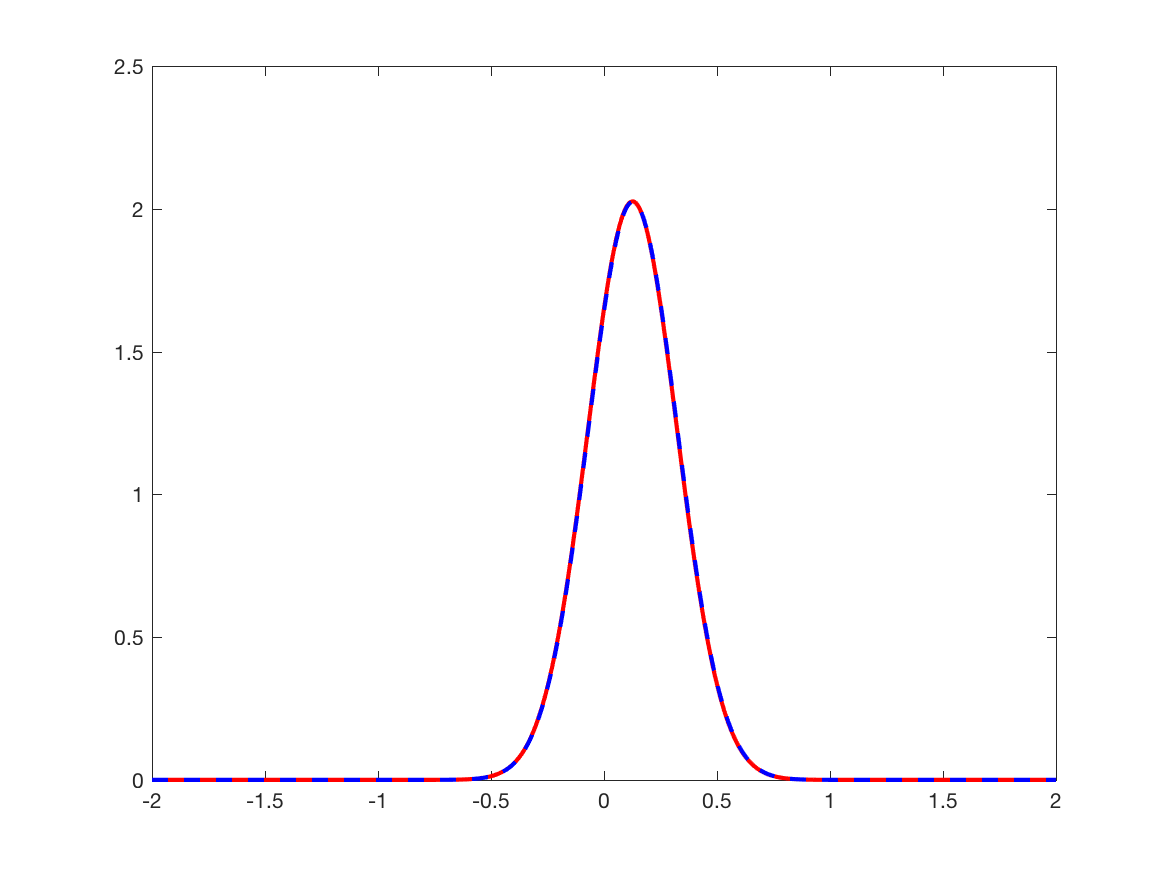}
        \caption{Time $t=100$.}
        \label{fig:lin4}
      \end{subfigure}
      \caption{Predicted state densities for different times. Kalman
        predicted (blue solid) and GMM predicted (red solid).}
      \label{fig:linear_example}
\end{figure}

To this end, consider the state-space model 
\begin{align}
  \label{eq:42}
  x_{t+1} &= \bmat{1 & 0.01 \\ 0 & 1} x_t + u_t + w_t,\\
  y_t &= \bmat{1 & 0} x_t + e_t, 
\end{align}
where the input was chosen as $u_t = \bmat{0 & 1}^T \bar{u}_t$ with $\bar{u}_t\sim \mathcal{N}(0,0.04)$ and
\begin{align}
  \label{eq:34}
  x_1 \sim \mathcal{N}(0, 1), \quad w_t \sim \mathcal{N}(0,
  0.01), \quad e_t \sim \mathcal{N}(0, 0.1),
\end{align}
We simulated $N=100$ samples from the system and then ran both a
Kalman Filter and the GMMF method according to
Algorithm~\ref{alg:GMMF}. The predicted mean of the first state is
shown in Figure~\ref{fig:linear_time}. Notice that the GMMF predicted
mean initially differs from the Kalman predicted mean. This was a
deliberate choice where the GMMF was initialised with an incorrect
prior, as depicted in Figure~\ref{fig:lin1}. This was generated by
choosing $25$ components in the prior that were evenly spaced between
$(-10,10)$ in both states. 

The purpose here is to demonstrate that the GMMF approach can quickly
converge to the correct number of modes and correct for the model
mismatch. Indeed, after just $t=7$ steps, the GMMF algorithm had
converged to one mixture component in both the predicted and
filtered mixtures. The convergence of the predicted PDFs can be
observed in Figure~\ref{fig:linear_example} as time progresses.

\subsection{Gaussian Mixture Model}
\begin{figure}[!bth]
  \includegraphics[width=\columnwidth]{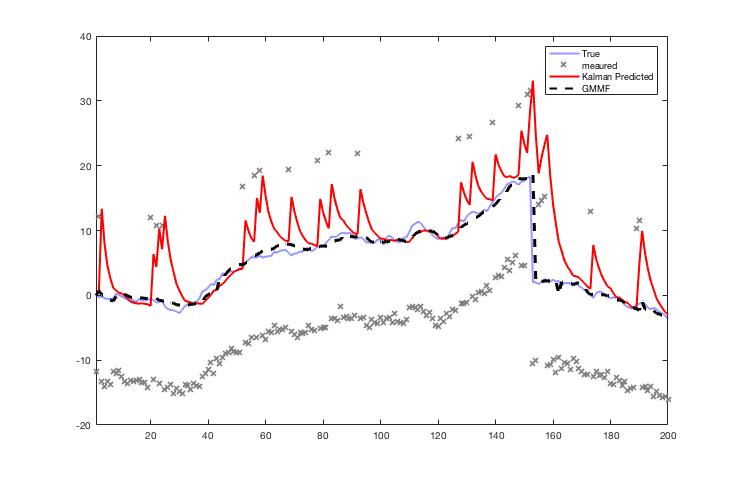}
  \caption{Plot of the predicted mean for the first state from both
    the Kalman Filter (red) and GMM Filter (dashed black) against the true
    state (solid blue). Measurements are indicated as crosses.}
  \label{fig:gmm_model_time}
\end{figure}

In this example, we are interested in profiling the GMMF algorithm for
a model that satisfies the assumptions \eqref{eq:5}--\eqref{eq:7}, and
yet is not easily amenable to Extended Kalman filtering or Unscented
Kalman filtering. Specifically, consider the following GMM state process
model with
{\small
\begin{align*}
  N_p &=1,& \mu_1 &= 0,& P_1 &= I_{2\times2},\\
  N_x &=2, & A^1 &= \bmat{1 & 0.1\\0 & 1}, & A^2 &= \bmat{0.1 & 0.01\\ 0 & 0.1},\\
  & &Q^1 &= 0.1^2I_{2\times2}, & Q^2 &= 0.003^2I_{2\times2},\\
  \beta^1 &= 0.99, & \beta^2 &= 0.01, & u_t^1 &= u_t^2 = \bmat{\sin(4\pi t/N) \\ 0},
\end{align*}
}%
and where the measurement model is described by
\begin{align*}
  N_y &=2, \quad C^1 = \bmat{1 & 0}, \quad C^2 = \bmat{1 & 0},\\
  v_t^1 &= 12.5,\quad v_t^2 = -12.5,\\
  R^1 &= 0.1, \quad R^2 = 0.1,\quad \gamma^1 = 0.1, \quad \gamma^2 = 0.9.
\end{align*}
We simulated $N=200$ samples from this system and used
the GMMF method to estimate the state
distribution. Figure~\ref{fig:gmm_model_time} shows the predicted mean
of the first state. For comparison, we also ran a Kalman filter
for the most likely model (according to the weight
$\beta^1, \beta^2, \gamma^1, \gamma^2$). Note that when the modelling
assumption is correct, e.g., from $t=100$ to $t=130$, the Kalman mean aligns very well with the GMMF mean as expected.

\subsection{Bi-modal Filtered Density Model} \label{sec:bimodal_example}
In this example, we demonstrate the GMMF algorithm on a model that,
due to its measurement equation, has a bi-modal filtered density.  The
prior and process models satisfy assumptions \eqref{eq:5} and
\eqref{eq:6}. Specifically, consider the following GMM state process
model:
\begin{align}
  N_p &= 50, \quad \mu_1^i = -10 + (i-1)\frac{20}{49}, \quad P_1^i = 0.1,\\
  N_x &= 1, \quad A_1 = 1, \quad u_t^1 = 5\cos(t_k),\\
  Q_1 &= 0.01, \quad \beta_1 = 1.
\end{align}
Here, however, we use a nonlinear measurement model that does not
strictly satisfy the structural assumption in
\eqref{eq:7}. Specifically, the measurement model is of the form
\begin{equation}
  y_k = h(x_t) + e_k, \quad h(x_t) = x_k^2, \quad e_k \sim \mathcal{N}(0,25).
\end{equation}
This particular measurement model was chosen to create
ambiguity in the filtered state between positive and negative values.

The nonlinear measurement model can be dealt with in the GMMF
framework by considering the Laplace approximation of the function for
each mixture component.  In this case, the measurement model for each
component becomes
\begin{align}
  C^{k,\ell}_t &=\frac{\partial h(x)}{\partial x}\Bigr|_{x = \hat{x}_{t|t-1}^\ell}, \quad v_t^{k,\ell} = h(\hat{x}_{t|t-1}^\ell) - C^{k,\ell}_t \hat{x}^\ell_{t|t-1}.
\end{align}
This approximates the filtered density in a similar way to a
first-order extended Kalman filter. In cases where the
mixture components have high variance, the approximation may be poor.
In this case, we are able to represent the prediction mixture by a
greater number of components than in \eqref{eq:9} by splitting
each component of the mixture into $N_s$ components.  Note that the
GMM model class easily accommodates this splitting process.

\begin{figure}[!bth]
  \includegraphics[width=\columnwidth]{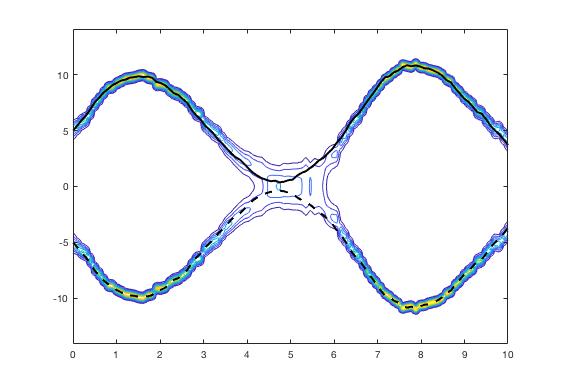}
  \caption{The filtered state density as predicted by the GMMF (contour), against the true state (solid black), and the true state mirrored about $x=0$ (dashed black).}
\label{fig:bimodal_filt_density}
\end{figure}

\begin{figure}[!bth]
    \centering
    \begin{subfigure}[t]{.4\linewidth}
        \includegraphics[width=\columnwidth]{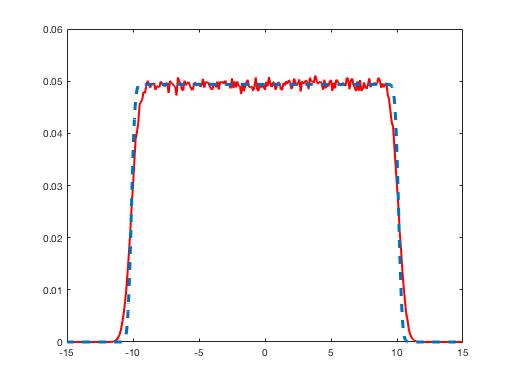}
       \caption{Predicated, Time $t=0.1$.}
        \label{fig:bm1}
      \end{subfigure}
          \begin{subfigure}[t]{.4\linewidth}
        \includegraphics[width=\columnwidth]{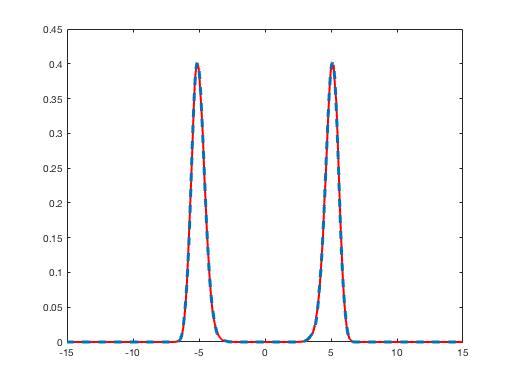}
       \caption{Filtered, Time $t=0.1$.}
        \label{fig:bm2}
      \end{subfigure}
      \\
      \begin{subfigure}[t]{.4\linewidth}
        \includegraphics[width=\columnwidth]{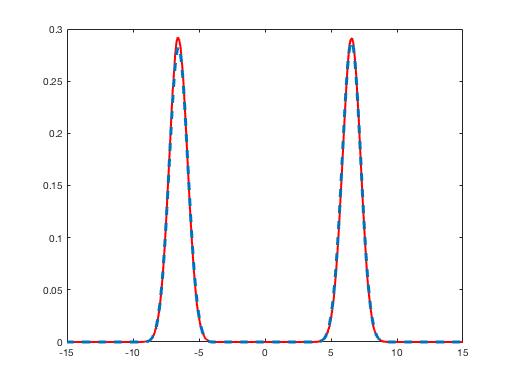}
        \caption{Predicted, Time $t=3$.}
        \label{fig:bm3}
      \end{subfigure}
        \begin{subfigure}[t]{.4\linewidth}
        \includegraphics[width=\columnwidth]{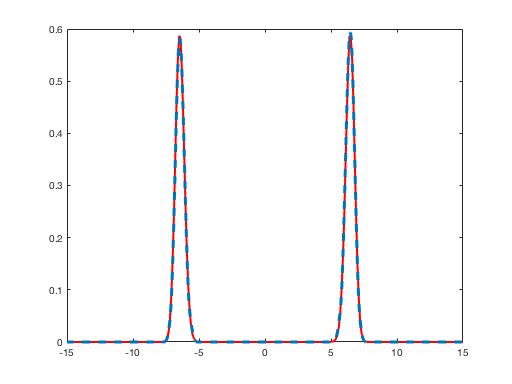}
        \caption{Filtered, Time $t=3$.}
        \label{fig:bm4}
      \end{subfigure}
      \\
      \begin{subfigure}[t]{.4\linewidth}
        \includegraphics[width=\columnwidth]{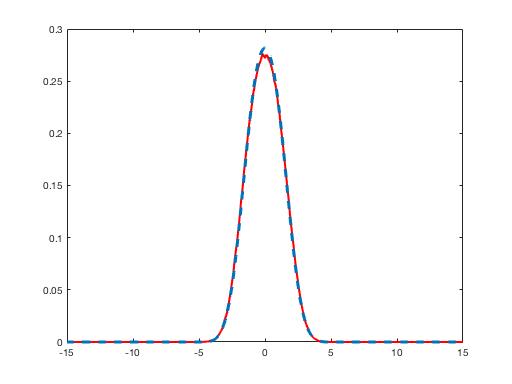}
       \caption{Predicted, Time $t=5$.}
        \label{fig:bm5}
      \end{subfigure}
        \begin{subfigure}[t]{.4\linewidth}
        \includegraphics[width=\columnwidth]{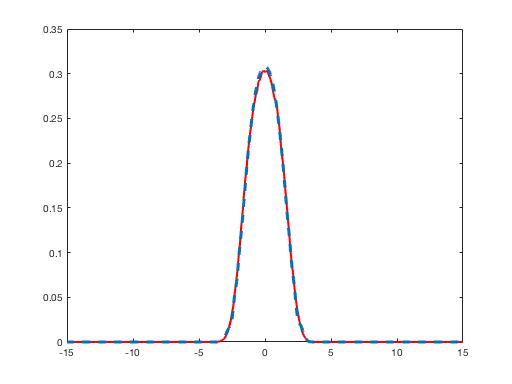}
       \caption{Filtered, Time $t=5$.}
        \label{fig:bm6}
      \end{subfigure}
      \\
      \begin{subfigure}[t]{.4\linewidth}
        \includegraphics[width=\columnwidth]{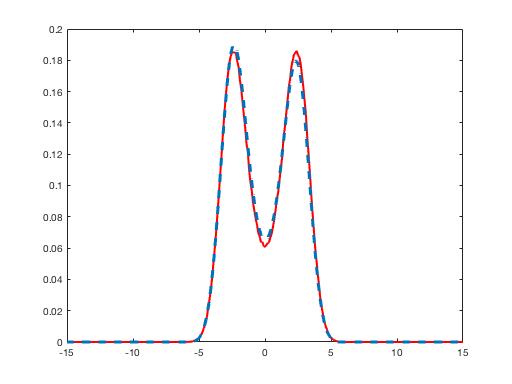}
        \caption{Predicted, Time $t=6$.}
        \label{fig:bm7}
      \end{subfigure}
        \begin{subfigure}[t]{.4\linewidth}
        \includegraphics[width=\columnwidth]{bimodal_k60_pred}
        \caption{Filtered, Time $t=6$.}
        \label{fig:bm8}
      \end{subfigure}
      \caption{The predicted and filtered state densities. In each case, the GMMF density (blue) is compared against the density from a SMC method (red).}
      \label{fig:bimodal}
\end{figure}

We simulated $N = 100$ input/output samples from this system, and ran
both the GMMF method and a Sequential Monte-Carlo (SMC) method to
estimate the state distribution. The filtered state densities from the
GMMF estimator are shown in Figure~\ref{fig:bimodal_filt_density}. The
predicted and filtered densities at various times are shown in
Figure~\ref{fig:bimodal}. The GMMF was initialised with a prior of
$50$ components evenly spaced between $(-10,10)$. 
Figure~\ref{fig:bm1} shows that the
GMMF and SMC generated densities are closely matched in this case,
despite the GMMF approach using no more than $35$ components in the
filtering mixture after the very first filtering step.

\subsection{Non-linear and Time Varying Model}
In this example, we demonstrate the GMMF algorithm on a model that
has both a nonlinear process and measurement model.  The purpose here
is to show that the GMMF filter has potential utlity for cases where
the model strictly violates the structural assumptions
\eqref{eq:5}--\eqref{eq:7}.  The chosen model has received significant
attention due it being recognised as a difficult state estimation
problem~\cite{DoucetGA:2000,GodsillDW:2004}.  The process and
measurement equations are given by
\begin{align*}
  x_{t+1} &= f(x_t) + w_t, \quad f(x_t) = ax_t + b\frac{x_t}{1+x_t^2}+c\cos(1.2t) \\
  y_t &= h(x_t) + v_t, \quad h(x_t) = dx_t^2,
\end{align*}
where $w_t \sim\mathcal{N}(0,1)$ and $e_t \sim \mathcal{N}(0,1)$, and the parameter values used are
\begin{align*}
  a = 0.5, \quad b = 25, \quad c = 8, \quad d = 0.05.
\end{align*}
As in Section~\ref{sec:bimodal_example}, the non-linearities are dealt
with by taking the Laplace approximations about the mean of each
mixture component via
\begin{align*}
    A^{j,s}_t &= \frac{\partial f(x)}{\partial x}\Bigr|_{ x = \hat{x}_{t|t}^s}, \quad u_t^{j,s} = f(\hat{x}_{t|t}^s) - A^{j,s}_t \hat{x}^s_{t|t}, \\
   C^{k,\ell}_t &=\frac{\partial h(x)}{\partial x}\Bigr|_{x = \hat{x}_{t|t-1}^\ell}, \quad v_t^{k,\ell} = h(\hat{x}_{t|t-1}^\ell) - C^{k,\ell}_t \hat{x}^\ell_{t|t-1}.
\end{align*}
In order to improve the approximation of the PDFs given by the
prediction and measurement steps, we represent the prediction and
filtering densities by a greater number of components than given by
\eqref{eq:9} and \eqref{eq:10}, respectively.  As in
Section~\ref{sec:bimodal_example}, this is achieved by splitting each
mixture component into $N_s $ components.

\begin{figure}[!bth]
    \centering
    \includegraphics[width=\columnwidth]{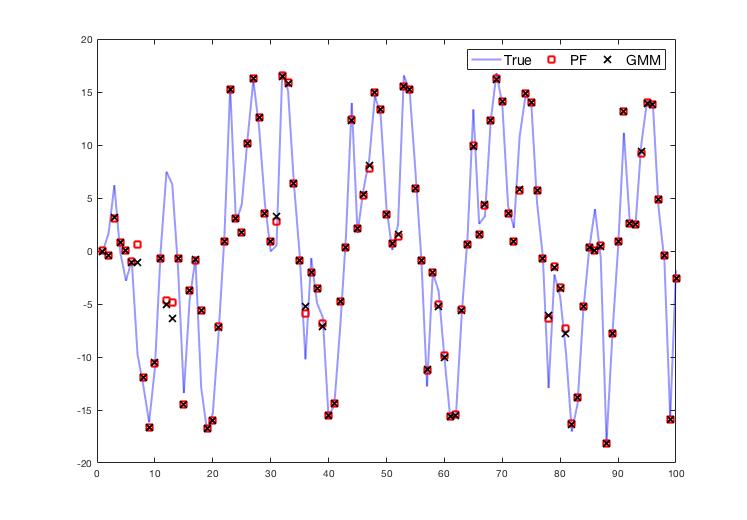}
    \caption{Plot of the predicted state mean from both the GMM Filter (red crosses) and the Particle filter (green squares) against the true state (solid blue).}
    \label{fig:nl_example}
\end{figure}

\begin{figure}[!bth]
    \centering
    \begin{subfigure}[t]{.4\linewidth}
        \includegraphics[width=\columnwidth]{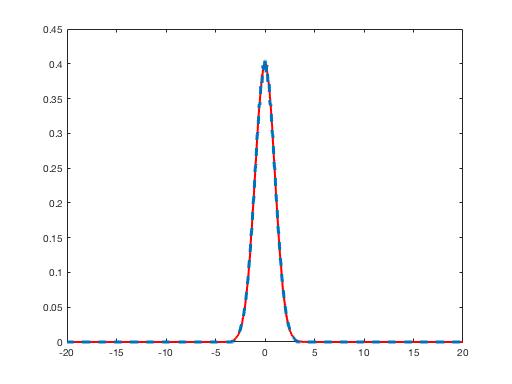}
       \caption{Predicated, Time $t=1$.}
        \label{fig:nl1}
      \end{subfigure}
          \begin{subfigure}[t]{.4\linewidth}
        \includegraphics[width=\columnwidth]{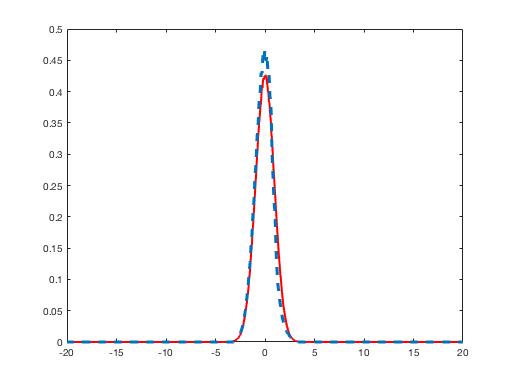}
       \caption{Filtered, Time $t=1$.}
        \label{fig:nl2}
      \end{subfigure}
      \\
      \begin{subfigure}[t]{.4\linewidth}
        \includegraphics[width=\columnwidth]{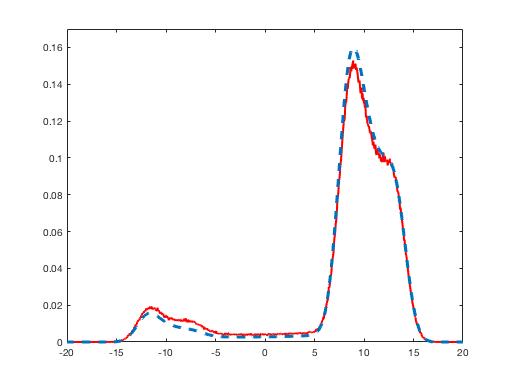}
        \caption{Predicted, Time $t=5$.}
        \label{fig:nl3}
      \end{subfigure}
        \begin{subfigure}[t]{.4\linewidth}
        \includegraphics[width=\columnwidth]{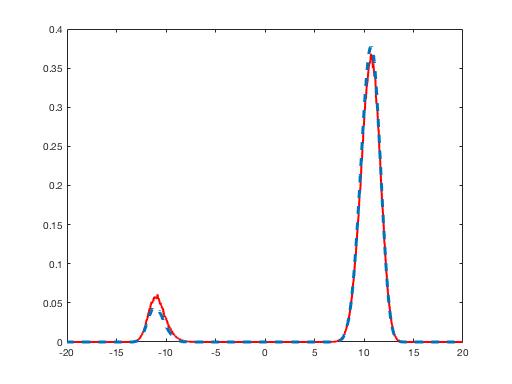}
        \caption{Filtered, Time $t=5$.}
        \label{fig:nl4}
      \end{subfigure}
      \\
      \begin{subfigure}[t]{.4\linewidth}
        \includegraphics[width=\columnwidth]{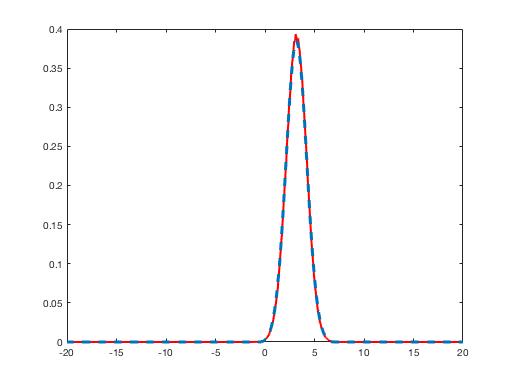}
       \caption{Predicted, Time $t=50$.}
        \label{fig:nl5}
      \end{subfigure}
        \begin{subfigure}[t]{.4\linewidth}
        \includegraphics[width=\columnwidth]{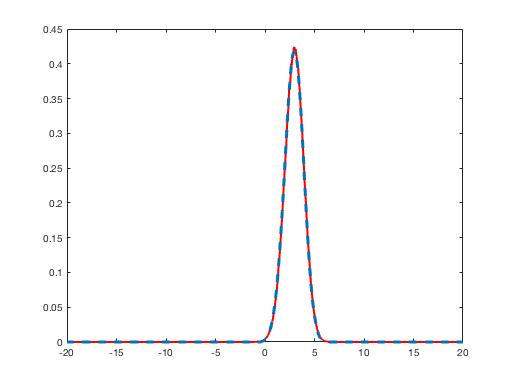}
       \caption{Filtered, Time $t=50$.}
        \label{fig:nl6}
      \end{subfigure}
      \\
      \begin{subfigure}[t]{.4\linewidth}
        \includegraphics[width=\columnwidth]{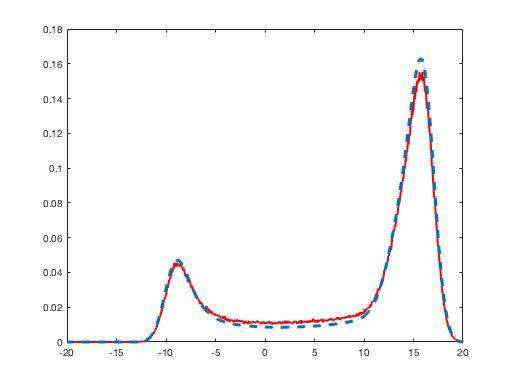}
        \caption{Predicted, Time $t=80$.}
        \label{fig:nl7}
      \end{subfigure}
        \begin{subfigure}[t]{.4\linewidth}
        \includegraphics[width=\columnwidth]{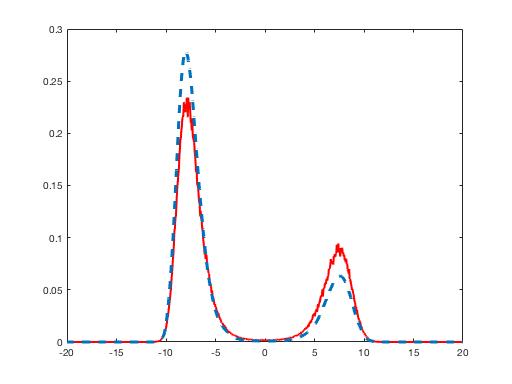}
        \caption{Filtered, Time $t=80$.}
        \label{fig:nl8}
      \end{subfigure}
      \caption{The predicted and filtered state densities. In each case, the GMMF density (blue) is compared against the density from a SMC method (red).}
      \label{fig:nl_example_densities}
\end{figure}

We simulated $N=100$ input/output samples from the system and ran both
the GMMF algorithm and, for comparison, a Sequential Monte-Carlo
method to estimate the state distributions.
Figure~\ref{fig:nl_example} shows the predicted mean of the state.
The predicted and filtered state densities are shown in
Figure~\ref{fig:nl_example_densities} for a selection of time
samples, where we note the close match between these
densities.

\section{CONCLUSIONS AND FUTURE WORKS}
In this paper we present a novel square-root form Bayesian filtering
algorithm for state-space models that can be described using a
Gaussian mixture for both the process and measurement PDFs. The main
attraction of restricting attention to this class of models is that the time and
measurement update equations can be solved in closed form. This comes
at the expense of exponential growth in computational load, which we
combat by employing a Kullback--Leibler reduction method at each stage
of the filter. The proposed algorithm is profiled on several examples
including non-linear models that fall strictly outside the model
class. Interestingly, this does not appear to pose a serious problem
for the algorithm.




\bibliographystyle{ieeetr}
\bibliography{brett,chris}

\end{document}